\documentclass[11pt]{article}

\usepackage[final]{acl}

\usepackage{times}
\usepackage{latexsym}

\usepackage[T1]{fontenc}

\usepackage[utf8]{inputenc}

\usepackage{microtype}

\usepackage{inconsolata}

\usepackage{graphicx}
\usepackage{amssymb}
\usepackage{amsmath}
\usepackage{makecell}
\usepackage{arydshln}
\usepackage{booktabs}
\usepackage{multirow}

\title{Rose-SQL: Role-State Evolution Guided Structured Reasoning for Multi-Turn Text-to-SQL}

\author{
    Le Zhou$^{1}$, Feng Yao$^{1}$, Fengcai Qiao$^{2}$, Bo Xu$^{3}$, 
    Fangyuan Wang$^{3}$\thanks{Corresponding authors},
    Boyan Xu$^{4}$\footnotemark[1] \\
    $^{1}$College of Systems Engineering, National University of Defense Technology \\
    $^{2}$College of Advanced Interdisciplinary Studies, National University of Defense Technology \\
    $^{3}$Institute of Automation, Chinese Academy of Sciences \\
    $^{4}$School of Computer Science, Guangdong University of Technology \\
    \texttt{\{zhoule116, yaofeng, fcqiao\}@nudt.edu.cn}\\
    \texttt{\{boxu,fangyuan.wang\}@ia.ac.cn} \quad
    \texttt{hpakyim@gmail.com}
}

\begin{document}
\maketitle
\maketitle
\begin{abstract}
Recent advances in Large Reasoning Models (LRMs) trained with Long Chain-of-Thought have demonstrated remarkable capabilities in code generation and mathematical reasoning. However, their potential in multi-turn Text-to-SQL tasks remains largely underexplored. Existing approaches typically rely on unstable API-based inference or require expensive fine-tuning on small-scale models. In this work, we present Rose-SQL, a training-free framework that leverages small-scale LRMs through in-context learning to enable accurate context-dependent parsing. We introduce the Role-State, a fine-grained representation that bridges the structural gap between schema linking and SQL generation by serving as a structural blueprint. To handle conversational dependencies, Rose-SQL traces the evolution of Role-State through historical context via structural isomorphism checks, guiding the model to infer the possible SQL composition for the current question through verified interaction trajectories. Experiments on the SParC and CoSQL benchmarks show that, within the Qwen3 series, Rose-SQL outperforms in-context learning baselines at the 4B scale and substantially surpasses state-of-the-art fine-tuned models at the 8B and 14B scales, while showing consistent gains on additional reasoning backbones.
\end{abstract}

\section{Introduction}
Text-to-SQL \cite{zhongSeq2SQL2017,xu2017sqlnet} is a fundamental semantic parsing task that translates natural language utterances into executable SQL queries over a given database schema, enabling users to access structured information without SQL expertise. While early research mainly focused on single-turn queries \cite{yu2018spider, bird}, real-world applications often involve multi-turn interactions \cite{yu2019sparc,yu2019cosql}, where users iteratively refine or build upon previous questions through dialogue. This setting requires the model to incorporate both the current utterance and relevant history to generate context-aware queries.

The rapid advancement of large language models (LLMs) has spurred new approaches to this task. In particular, LLMs enable strong in-context reasoning without explicit task-specific training, offering a flexible alternative to traditional supervised models. This has led to two dominant paradigms in multi-turn Text-to-SQL tasks: \textbf{in-context learning} (ICL) with large LLMs, and \textbf{fine-tuned} smaller models with customized architectures.
In-context learning approaches leverage proprietary LLMs such as GPT-3.5 to generate SQL queries from dialogue histories directly. Representative methods include ACT-SQL \cite{ACT-SQL}, which rewrites dependent questions and applies chain-of-thought (CoT) prompting \cite{CoT}, and CoE-SQL \cite{CoE-SQL}, which models SQL transitions to maintain consistency across turns. While flexible, these methods are highly sensitive to prompt quality and struggle with noisy or ambiguous historical context.
On the other hand, fine-tuned approaches use smaller open-source models with task-specific modules for schema linking, intent tracking, or SQL history modeling. For example, HIE-SQL \cite{HIE-SQL}, RASAT \cite{RASAT}, and Track-SQL \cite{Track-SQL} encode context using pretraining objectives or graph structures. While effective, these methods are training-intensive and costly to adapt.

Given these trade-offs, a natural question arises: \textit{Can small-scale open-source models perform effective multi-turn reasoning purely through in-context learning, without fine-tuning?}
We find that recent small-scale reasoning models—such as the Qwen3 \cite{Qwen3} and DeepSeek \cite{DeepSeek-R1} series—have made notable progress in in-context tasks, thanks to improved instruction tuning and long-chain-of-thought capabilities. These models now show strong reasoning ability, even without fine-tuning. Yet, their application to multi-turn Text-to-SQL remains underexplored. The main bottleneck is not model capacity, but the lack of mechanisms to manage evolving dialogue context and align schema linking with SQL structure. In the absence of strong inductive biases, small models often misinterpret context dependencies or fail to capture compositional query semantics.

\begin{figure}[t]
    \centering
    \includegraphics[width=1\linewidth]{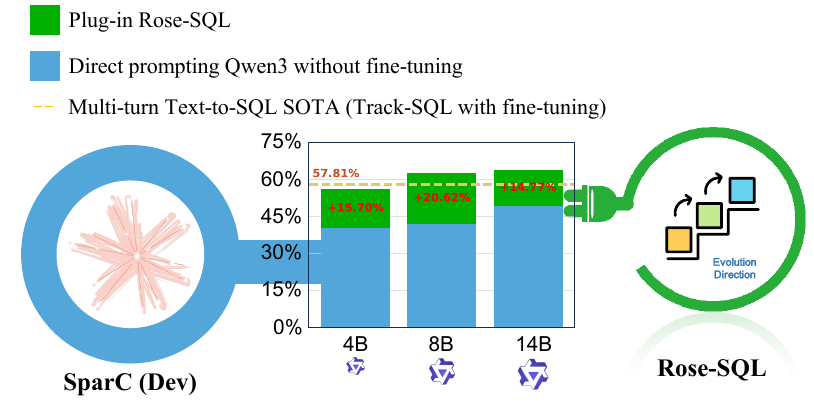}
    \caption{Rose-SQL: A plug-in framework boosting small open-source LLMs. Our method significantly improves Qwen3's performance by 14.77\%-20.62\% across different model sizes (4B-14B) without fine-tuning, surpassing even fine-tune based multi-turn Text-to-SQL SOTA (Track-SQL) on the SparC (Dev) dataset.}
    \label{fig:effectiveness}
\end{figure}

To bridge this gap, we propose \textbf{Rose-SQL}, a reasoning framework designed to adapt small-scale LRMs for context-dependent parsing without task-specific fine-tuning. As illustrated in Figure \ref{fig:effectiveness}, Rose-SQL facilitates the efficient utilization of historical context while achieving performance comparable to, or even surpassing, state-of-the-art fine-tuned methods. The cornerstone of our approach is the introduction of \textbf{Role-State}, a fine-grained schema linking representation that explicitly maps natural language entities to their specific functional roles within a SQL query. Serving as a structural blueprint, Role-State achieves the decoupling of semantic intent from syntactic execution, imposing rigid constraints that guide the model through a stabilized reasoning path. 

To manage complex dialogue dynamics, Rose-SQL operates through three synergistic phases. First, the \textbf{Gain Dependency Analysis} phase quantifies the contextual influence of historical utterances, enabling the framework to distinguish substantive information gain from irrelevant conversational noise. Subsequently, the \textbf{Evolutionary Trajectory Searching} mechanism identifies historical interaction patterns and verifies them via structural isomorphism checks on Role-State transitions, providing a validated reference to infer the possible SQL composition for the current question. Finally, the \textbf{Augmented Hierarchical Reasoning} stage integrates verified trajectories with contextual anchors, guiding the model to sequentially infer the intermediate thought process and Role-State before generating the final SQL code. Experimental results on the SParC and CoSQL benchmarks demonstrate that Rose-SQL significantly enhances SQL generation accuracy across various parameter scales.

Our contributions are summarized as follows:
\begin{itemize}
    \item We propose the Rose-SQL framework, which leverages the Role-State mechanism to decouple semantic mapping from syntactic generation, providing a structured intermediate target for small-scale LRMs.
    \item We devise an evolutionary guidance strategy that integrates Gain Dependency Analysis with structural isomorphism checks, effectively mitigating contextual noise by verifying the logical consistency of dialogue evolution to infer the possible SQL composition for the current question.
    \item We introduce an augmented hierarchical reasoning paradigm that stabilizes the deduction process via verified trajectories and contextual anchors, achieving state-of-the-art performance on the validation sets of two benchmarks SparC and CoSQL.
\end{itemize}

\section{Methodology}

In the context of multi-turn Text-to-SQL parsing, the task is formulated as mapping a sequence of conversational utterances $\mathcal{Q}_{\le i}$ and the corresponding database schema $\mathcal{S} = (t_m, c_{m, n_m})$ to a target SQL query $s_m$. Here, $t_m$ denotes the $m^{\text{th}}$ table in the database, while $c_{m, n_m}$ represents the $n_m^{\text{th}}$ column within table $t_m$. In this section, we will provide an overview of the framework designed to solve this problem and delve into its design details.

\begin{figure}[htbp]
    \centering
    \includegraphics[width=0.49\textwidth]{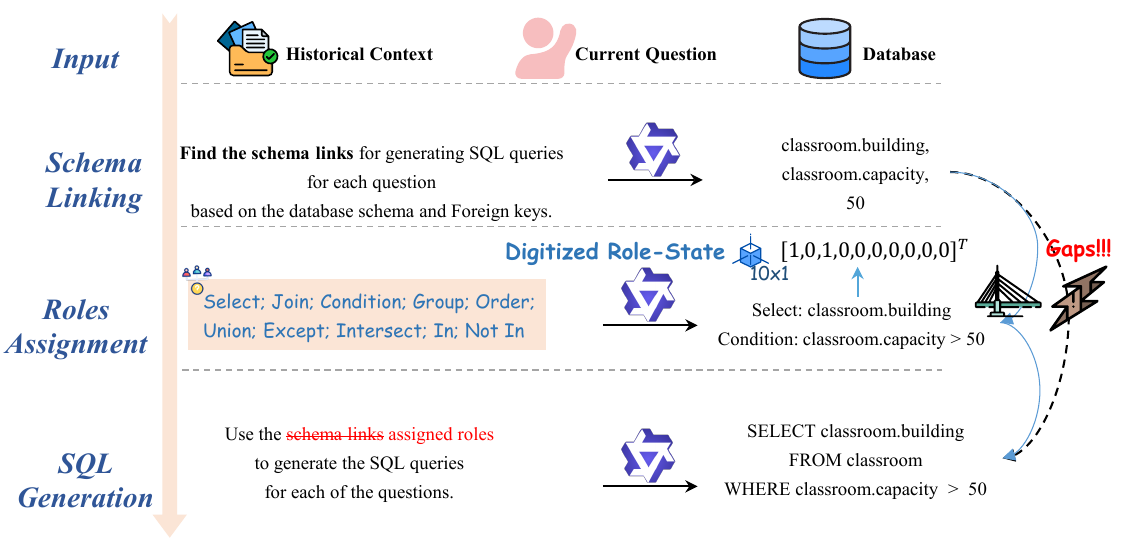}
    \caption{The Role-State mechanism: A structural blueprint bridging schema linking and SQL generation.}
    \label{fig:role-state}
\end{figure}

\begin{figure*}[t]
\centering
\includegraphics[width=0.96\textwidth]{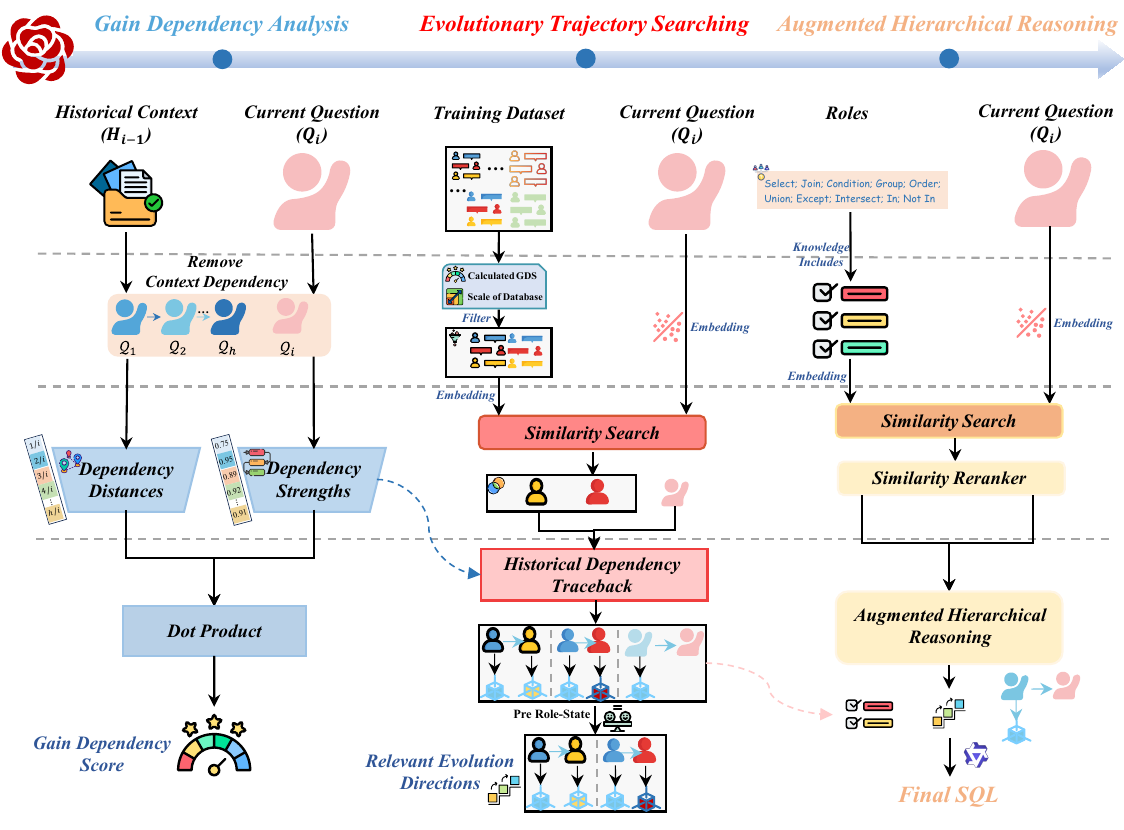} %
\caption{An overview of our framework. Rose-SQL first performs a context-aware dependency quantification via Gain Dependency Analysis to distinguish substantive information gain from conversational noise. It then conducts an isomorphism-guided evolutionary search to verify historical patterns through digitized Role-State transitions, ensuring retrieved trajectories are logically analogous to the current dialogue. Finally, an augmented hierarchical reasoning stage utilizes these verified trajectories to infer the possible SQL composition for the current question, guiding the model to sequentially produce the Role-State blueprint and final executable query.}
\label{Rose-SQL}
\end{figure*}

\subsection{Overview of Rose-SQL}
To address structural misalignment and contextual noise in multi-turn Text-to-SQL parsing, we propose Rose-SQL, a training-free framework built upon hierarchical structural guidance. At the core of Rose-SQL is the Role-State, an intermediate structural blueprint that decouples semantic intent from syntactic realization by mapping relevant entities to their functional roles in SQL composition. These role assignments are further encoded as a ten-dimensional indicator vector, enabling explicit structural comparison across dialogue turns and providing a computable foundation for subsequent structural verification.

Building upon this representation, Rose-SQL verifies evolutionary trajectories from historical interactions to identify logically analogous transitions, so that historical context can be transformed from raw dialogue memory into reliable structural evidence for the current turn. The verified trajectories, together with contextual anchors and role-level knowledge, are then integrated into an augmented hierarchical reasoning process that guides the model to first infer the intermediate structural plan and then generate the final executable SQL. In this way, Rose-SQL provides a more stable path from conversational intent to formal query generation. The overall architecture of the framework is illustrated in Figure~\ref{Rose-SQL}.

\subsection{Role-State Definition and Digitization}
To address structural ambiguity in context-dependent parsing, we introduce the \textit{Role-State}, a unified representation that serves as a structural blueprint for the target SQL query. As illustrated in Figure~\ref{fig:role-state}, the Role-State is constructed by assigning functional roles to the schema elements involved in the current query and encoding the resulting global query structure as a ten-dimensional indicator vector $\mathbf{v} \in \{0,1\}^{10}$. Specifically, the vector records the presence of the following structural roles: \textit{selected, join, condition, order, group, union, except, intersect, in}, and \textit{nin}. 

This digitized representation provides a computable basis for structural isomorphism checks across dialogue turns. By converting structural transitions into explicit and comparable signals, Rose-SQL can determine whether a historical transition is logically analogous to the current interaction, thereby inferring the possible SQL composition of the current question. In this way, the Role-State not only preserves structural consistency throughout dialogue evolution, but also supports a more stable progression from semantic intent to executable SQL generation.

\subsection{Gain Dependency Analysis}

The Gain Dependency Analysis phase quantifies how strongly the current question depends on its dialogue history, with the goal of distinguishing substantive contextual gain from irrelevant conversational noise. To obtain a more precise estimate, we first rewrite each question using the M-schema \cite{XiYanSQL} to reduce context dependence, thereby enabling a cleaner assessment of how much information is contributed by previous turns. At turn $i$, the dependency between the rewritten question $\mathcal{Q}_i$ and its preceding history is modeled by two vectors of length $i-1$.

The Dependency Strengths vector $\mathbf{S}_i \in \mathbb{R}^{i-1}$ measures the semantic gain brought by each historical turn $h \in \{1, \dots, i-1\}$ to $\mathcal{Q}_i$ through a normalized perplexity reduction:
\begin{equation}
    s_{h,i} = \frac{PPL(\mathcal{Q}_i) - PPL(\mathcal{Q}_i \mid \mathcal{Q}_h)}{PPL(\mathcal{Q}_i)}
    \label{eq:gdst}
\end{equation}
where $PPL(\mathcal{Q}_i)$ and $PPL(\mathcal{Q}_i \mid \mathcal{Q}_h)$ denote the standalone and conditional perplexities, respectively. Intuitively, a larger $s_{h,i}$ indicates that the historical turn $\mathcal{Q}_h$ provides more useful information for interpreting the current question.

Meanwhile, the Dependency Distances vector $\mathbf{D}_i \in \mathbb{R}^{i-1}$ captures temporal proximity by assigning
\begin{equation}
    d_{h,i} = \frac{h}{i},
\end{equation}
which assigns larger weights to more recent turns. As illustrated in Figure~\ref{Rose-SQL}, the final Gain Dependency Score is computed as
\begin{equation}
    GDS_i = \mathbf{S}_i \cdot \mathbf{D}_i.
\end{equation}
This score provides a compact characterization of the dependency footprint of the current question and serves as the primary filtering signal for retrieving structurally analogous interaction trajectories. In this way, subsequent Role-State isomorphism checks can be focused on historically relevant and structurally plausible candidates.

\subsection{Evolutionary Trajectories Searching}
The Evolutionary Trajectories Searching phase identifies historical interaction patterns from the training set that are both semantically relevant and structurally analogous to the current dialogue. As illustrated in Figure~\ref{Rose-SQL}, this phase follows a dual-stage filtering procedure. We first retrieve candidate training dialogues $\mathcal{D}_m$ whose dependency footprint and schema scale---including the numbers of tables, columns, and foreign keys---are compatible with the current interaction. Within each candidate dialogue, we then locate the turn that is most semantically similar to the current question $\mathcal{Q}_i$, together with its strongest historical dependency, as follows:
\begin{equation}
    idx = \mathop{\arg\max}_{j} \mathrm{Sim}(\mathcal{Q}_i, \mathcal{Q}^{train}_{m,j}), \quad j \in \{1, \dots, L\}
\end{equation}
\begin{equation}
    idx_{dep} = \mathop{\arg\max}_{j} GDS_{j,idx}, \quad j \in \{1, \dots, idx-1\}
\end{equation}
where $GDS_{j,idx}$ denotes the dependency score of turn $idx$ on turn $j$ in the retrieved training exemplar. Correspondingly, for the current dialogue, we denote by $\mathcal{Q}_{i_{dep}}$ the strongest historical dependency of the current question $\mathcal{Q}_i$ under the same Gain Dependency Analysis.

The final selection is governed by the Role-State Isomorphism Check. Since Role-States are precomputed for all training samples, we retain a candidate transition only when the digitized Role-State of the training anchor $idx_{dep}$ is isomorphic to that of the current contextual anchor $\mathcal{Q}_{i_{dep}}$; otherwise, the candidate is discarded. If this condition is satisfied, the transition is treated as a valid evolutionary trajectory $\mathcal{P}_i$, and the corresponding hierarchical reasoning trace $\mathcal{R}_{idx}$ is adopted as a structural reference. This verified guidance allows the model to infer the possible SQL composition of the current question by following a logically analogous evolution path.

Specifically, each valid trajectory is represented as a quadruple:
\begin{equation}
    \mathcal{P}_{i} = \{ \mathcal{Q}_{idx_{dep}}^{train}, \mathcal{R}_{idx_{dep}}, \mathcal{Q}_{idx}^{train}, \mathcal{R}_{idx} \}
\end{equation}
where $\mathcal{R}$ denotes the hierarchical reasoning trace grounded by the corresponding Role-State. The first pair records the historical anchor transition, while the second pair provides the aligned target transition that serves as structural guidance for the current turn. To balance retrieval reliability and prompt efficiency, we retain only the single most significant historical transition in each trajectory, ensuring that the model receives high-confidence structural evidence without introducing excessive contextual ambiguity.

\subsection{Augmented Hierarchical Reasoning}

The final stage of Rose-SQL integrates structural and contextual signals to guide the large reasoning model through a stabilized deduction process. As illustrated in Figure~\ref{Rose-SQL}, the framework formulates the predicted SQL query $\mathcal{Y}_i^{pred}$ by conditioning on the task instruction $\mathcal{I}$, database schema $\mathcal{S}$, current question $\mathcal{Q}_i$, and augmented contextual evidence:
\begin{equation}
    \mathcal{Y}_i^{pred} = LRM(\mathcal{I}, \mathcal{S}, \mathcal{Q}_i, \mathcal{C}_{ctx}, \{\mathcal{P}_i\}_{i=1}^{N}, \mathcal{E}_{rki})
    \label{eq:final_task}
\end{equation}
Here, the prompt incorporates the Contextual Anchor $\mathcal{C}_{ctx}$ to focus the model on the most relevant historical context, verified Evolutionary Trajectories $\{\mathcal{P}_i\}_{i=1}^{N}$ to provide structurally analogous transition guidance, and Role Knowledge Injection $\mathcal{E}_{rki}$ to supply atomic knowledge of SQL functional roles (details in Appendix~\ref{rki}). 

Rather than directly generating SQL from the natural language question, Rose-SQL adopts a hierarchical reasoning process in which the model first infers the intermediate structural plan and Role-State, and then produces the final executable query. This design reduces the structural gap between semantic understanding and SQL realization, helping the model maintain conversational coherence and structural consistency throughout generation. As a result, Rose-SQL provides a more stable path from multi-turn intent interpretation to executable SQL without task-specific fine-tuning.

\begin{table*}[t]
    \centering
    \setlength{\tabcolsep}{0.8mm} 
    \renewcommand{\arraystretch}{1.1}

    \resizebox{\textwidth}{!}{%
    \begin{tabular}{c|ccc|ccc|ccc|ccc}
        \hline
        \multirowcell{3}{\textbf{Model}} & \multicolumn{6}{c|}{\textbf{SparC}} & \multicolumn{6}{c}{\textbf{CoSQL}} \\
        \cline{2-13}
        & \multicolumn{3}{c|}{\textbf{QM}} & \multicolumn{3}{c|}{\textbf{IM}} & \multicolumn{3}{c|}{\textbf{QM}} & \multicolumn{3}{c}{\textbf{IM}} \\
        \cline{2-13}
        & \textbf{EM $\uparrow$} & \textbf{EX $\uparrow$} & \textbf{TS $\uparrow$} & \textbf{EM $\uparrow$} & \textbf{EX $\uparrow$} & \textbf{TS $\uparrow$} & \textbf{EM $\uparrow$} & \textbf{EX $\uparrow$} & \textbf{TS $\uparrow$} & \textbf{EM $\uparrow$} & \textbf{EX $\uparrow$} & \textbf{TS $\uparrow$} \\
        \hline
        
        \multicolumn{13}{c}{\textbf{\textit{In-Context Learning Approach}}} \\
        \hline
        ACT-SQL \cite{ACT-SQL} & 51.0 & 63.8 & 56.9 & 24.4 & 38.9 & 29.6 & 46.0 & 63.7 & 55.2 & 13.3 & 30.7 & 21.5 \\
        CoE-SQL \cite{CoE-SQL} & 56.0 & 70.3 & 63.3 & 36.5 & 50.5 & 41.9 & 52.4 & 69.6 & 60.6 & 23.9 & 39.6 & 30.4 \\
        \hline

        \multicolumn{13}{c}{\textbf{\textit{Fine-tuned Model}}} \\
        \hline
        HIE-SQL + GraPPa \cite{HIE-SQL} & 64.7 & - & - & 45.0 & - & - & 56.4 & - & - & 28.7 & - & - \\
        RASAT + PICARD \cite{RASAT} & 67.7 & 73.3 & - & 49.1 & 54.0 & - & 58.8 & 67.0 & - & 27.0 & 39.6 & - \\
        QDA-SQL \cite{QDA-SQL} & 61.3 & - & - & 44.1 & - & - & 57.3 & - & - & 30.0 & - & - \\
        Track-SQL \cite{Track-SQL} & 65.41 & 75.39 & 69.16 & 46.91 & 57.81 & 50.71 & 58.19 & 71.10 & 65.54 & 28.67 & 45.05 & 36.17 \\
        \hline

        \multicolumn{13}{c}{\textbf{\textit{Ours}}} \\
        \hline
        Qwen3-4B & 54.11 & 66.00 & 58.27 & 28.67 & 40.28 & 31.04 & 54.20 & 62.07 & 56.11 & 27.30 & 31.06 & 25.25 \\
        + Rose-SQL & 61.51 & 74.23 & 68.74 & 43.83 & 55.98 & 48.95 & 55.61 & 72.39 & 66.27 & 29.01 & 45.39 & 38.57 \\
        \cdashline{1-13}
        Qwen3-8B & 53.12 & 66.99 & 61.68 & 26.78 & 41.94 & 35.31 & 56.50 & 65.32 & 60.66 & 25.94 & 34.57 & 30.38 \\
        + Rose-SQL & 65.75 & 78.72 & 72.73 & 50.00 & 62.56 & 55.21 & 63.16 & 74.48 & 68.62 & 36.18 & 50.85 & 41.30 \\
        \cdashline{1-13}
        Qwen3-14B & 55.19 & 71.98 & 65.25 & 29.38 & 49.05 & 40.52 & 58.78 & 70.19 & 63.14 & 27.64 & 42.32 & 32.76 \\
        + Rose-SQL & 66.45 & 79.10 & 74.44 & 50.25 & 63.82 & 58.04 & 62.26 & 77.36 & 72.19 & \textbf{37.88} & \textbf{53.92} & \textbf{47.09} \\
        \cdashline{1-13}
        DeepSeek-R1-Distill-Llama-8B \cite{DeepSeek-R1} & 31.58 & 49.13 & 43.89 & 10.43 & 21.09 & 16.82 & 32.94 & 47.36 & 42.62 & 9.95 & 17.27 & 13.91 \\
        + Rose-SQL & 42.50 & 61.25 & 54.30 & 18.50 & 32.80 & 25.40 & 39.70 & 59.17 & 52.07 & 13.19 & 26.99 & 19.86 \\
        \cdashline{1-13}
        gpt-oss-20b \cite{openai2025gptoss120bgptoss20bmodel} & 60.10 & 73.50 & 66.80 & 33.20 & 50.80 & 36.50 & 62.68 & 70.85 & 64.86 & 31.67 & 41.61 & 30.19 \\
        + Rose-SQL & \textbf{68.50} & \textbf{80.25} & \textbf{75.30} & \textbf{52.40} & \textbf{65.40} & \textbf{59.50} & \textbf{63.98} & \textbf{77.52} & \textbf{72.21} & 37.36 & 53.82 & 46.53 \\
        \hline
    \end{tabular}%
    }
    \caption{Overall performance comparison on SparC and CoSQL development sets. The best confirmed results are \textbf{bolded}.}
    \label{tab:main_results}
\end{table*}

\section{Experiment Setup}

\paragraph{Models.} 
To evaluate the efficacy of the Rose-SQL framework, we primarily utilize the Qwen3 series models, with Qwen3-8B serving as our main experimental subject. These models are selected as a representative testbed for small-scale LRMs due to their advanced long-chain-of-thought capabilities. By conducting evaluations across 4B, 8B, and 14B parameter scales, we comprehensively assess the effectiveness, robustness, and scalability of our proposed approach. In addition, to further examine whether the gains of Rose-SQL transfer beyond a single backbone family, we also evaluate two additional reasoning models, namely DeepSeek-R1-Distill-Llama-8B \cite{DeepSeek-R1} and gpt-oss-20b \cite{openai2025gptoss120bgptoss20bmodel}, using the same Rose-SQL pipeline and evaluation protocol.
\paragraph{Hyperparameters.}
We perform inference using the vLLM framework with a greedy decoding strategy and a sampling temperature of 0 to ensure deterministic, syntactically valid SQL generation. The in-context learning setup follows a fixed-shot configuration, incorporating two Role Knowledge Injection examples per role type. For Evolutionary Trajectories, we provide two candidates per question for the SParC dataset and one for CoSQL. This configuration is calibrated to balance supervisory diversity with context length constraints, ensuring a focused and efficient inference process. Unless otherwise specified, all compared reasoning backbones are evaluated under the same decoding and prompting protocol. The additional inference cost of Rose-SQL mainly stems from the increased prompt length introduced by retrieved trajectories and role exemplars, while generation itself remains a single deterministic decoding pass.

\paragraph{Datasets.} We evaluate our proposed method on the SParC \cite{yu2019sparc} and CoSQL \cite{yu2019cosql} benchmarks. SParC comprises 4,298 coherent question sequences, encompassing over 12k individual questions and their corresponding SQL queries. CoSQL features more than 10k annotated SQL queries, where each dialogue is designed to simulate a real-world scenario: an interaction between a non-expert user exploring a database and an expert formulating SQL to retrieve the requested information. Furthermore, the dataset includes an evaluation script that divides SQL queries into four difficulty levels.

\paragraph{Evaluation metrics.}
To evaluate the performance of our method in the Text-to-SQL task, we employ two official metrics, namely Question Match (QM) and Interaction Match (IM), in addition to three commonly adopted sub-metrics: Exact Match Accuracy (EM), Execution Accuracy (EX), and Test Suite Accuracy (TS). The QM metric measures whether the predicted SQL query correctly corresponds to a single question, whereas IM metric determines whether all predicted queries within a multi-turn interaction satisfy the QM criterion. Specifically, the EM metric evaluates the structural correctness of the predicted SQL query; The EX metric focuses on whether the execution result of the predicted SQL matches the ground truth; TS is a stricter metric that not only checks execution correctness, but also requires that the predicted query yields the correct results across multiple database instances and schemas.

\section{Main Results}
As presented in Table \ref{tab:main_results}, experimental results demonstrate that Rose-SQL significantly elevates performance across all evaluated parameter scales. Notably, our framework enables the Qwen3-4B model to achieve performance parity with the 8B-scale baseline, effectively bridging the gap between model capacities through structural guidance. On the SParC dataset, Rose-SQL yields a substantial increase in EX by 11.73\% and TS by 11.05\% for single-turn queries; for multi-turn interactions, EX and TS improve by 6.96\% and 7.56\%, respectively. Consistent trends are observed on the CoSQL benchmark, underscoring the generality of our reasoning-enhanced approach. These gains emphasize that integrating structural Role-State evolution significantly strengthens the multi-turn Text-to-SQL capabilities of small-scale models.

Within the Qwen3 series, Rose-SQL also compares favorably with both prompting-based and fine-tuned baselines. In particular, the Rose-SQL-8B configuration surpasses GPT-3.5-turbo-based methods such as ACT-SQL and CoE-SQL on both SParC and CoSQL under the training-free setting. Furthermore, Rose-SQL demonstrates strong robustness relative to fine-tuned approaches. While RASAT+PICARD relies on SQL-specific post-processing and Track-SQL requires auxiliary context-tracking modules, Rose-SQL attains competitive or superior multi-turn accuracy without task-specific fine-tuning or secondary correction strategies. This trend is also consistent with the more detailed comparison against SFT baselines in Appendix \ref{Ablation Studies}, where weaker base reasoning models may lag behind task-specific experts, while structural guidance substantially narrows or even reverses this gap.

In addition to the Qwen3 series, we further evaluate Rose-SQL on two additional reasoning backbones, namely DeepSeek-R1-Distill-Llama-8B \cite{DeepSeek-R1} and gpt-oss-20b \cite{openai2025gptoss120bgptoss20bmodel}, under the same Rose-SQL pipeline and evaluation protocol. Rose-SQL consistently improves both question-level and interaction-level performance over the corresponding base models. Notably, although DeepSeek-R1-Distill-Llama-8B remains weaker than the strongest competitors in absolute terms, this mainly reflects its substantially lower starting point in our evaluation setting: on SParC, Rose-SQL improves its QM-EX from 49.13\% to 61.25\% and its IM-EX from 21.09\% to 32.80\%. In contrast, when applied to the stronger gpt-oss-20b backbone, Rose-SQL further reaches 80.25\% QM-EX and 65.40\% IM-EX on SParC. These results indicate that Rose-SQL provides transferable structural gains across different reasoning backbones, while the final performance ceiling still depends on the strength of the underlying base model.

\begin{table}[h]
    \centering
    \setlength{\tabcolsep}{1.2mm}
    \resizebox{\columnwidth}{!}{ 
    \begin{tabular}{l ccc ccc}
        \toprule
        & \multicolumn{3}{c}{\textbf{QM}} & \multicolumn{3}{c}{\textbf{IM}} \\
        \cmidrule(lr){2-4} \cmidrule(lr){5-7}
        \textbf{Method} & \textbf{EM $\uparrow$} & \textbf{EX $\uparrow$} & \textbf{TS $\uparrow$} & \textbf{EM $\uparrow$} & \textbf{EX $\uparrow$} & \textbf{TS $\uparrow$} \\
        \midrule
        Qwen3-8B (Base) & 53.12 & 66.99 & 61.68 & 26.78 & 41.94 & 35.31 \\
        + Role-State       & 53.37 & 72.57 & 65.37 & 30.80 & 48.57 & 39.57 \\
        + $\mathcal{E}_{rki}$  & 60.43 & 71.82 & 65.75 & 39.73 & 49.52 & 41.94 \\
        + $\mathcal{P}_i$      & 63.76 & 76.16 & 70.00 & 44.73 & 59.87 & 53.89 \\
        \midrule
        \textbf{+ $\mathcal{C}_{ctx}$ (Rose-SQL)} & \textbf{65.75} & \textbf{78.72} & \textbf{72.73} & \textbf{50.00} & \textbf{62.56} & \textbf{55.21} \\
        \bottomrule
    \end{tabular}
    }
    \caption{Ablation results of Rose-SQL on the SParC dev set, illustrating the cumulative performance gains.}
    \label{tab:ablation_sparc}
\end{table}

\begin{table}[h]
    \centering
    \setlength{\tabcolsep}{1.2mm}
    \resizebox{\columnwidth}{!}{ 
    \begin{tabular}{l ccc ccc}
        \toprule
        & \multicolumn{3}{c}{\textbf{QM}} & \multicolumn{3}{c}{\textbf{IM}} \\
        \cmidrule(lr){2-4} \cmidrule(lr){5-7}
        \textbf{Method} & \textbf{EM $\uparrow$} & \textbf{EX $\uparrow$} & \textbf{TS $\uparrow$} & \textbf{EM $\uparrow$} & \textbf{EX $\uparrow$} & \textbf{TS $\uparrow$} \\
        \midrule
        Qwen3-8B (Base) & 56.50 & 65.32 & 60.66 & 25.94 & 34.57 & 30.38 \\
        + Role-State       & 57.56 & 72.17 & 66.07 & 24.48 & 40.00 & 31.72 \\
        + $\mathcal{E}_{rki}$  & 58.49 & 72.69 & 65.54 & 30.03 & 44.03 & 34.81 \\
        + $\mathcal{P}_i$      & 62.58 & 73.27 & 69.92 & 33.40 & 49.44 & 39.20 \\
        \midrule
        \textbf{+ $\mathcal{C}_{ctx}$ (Rose-SQL)}  & \textbf{63.16} & \textbf{74.48} & \textbf{68.62} & \textbf{36.18} & \textbf{50.85} & \textbf{41.30} \\
        \bottomrule
    \end{tabular}
    }
    \caption{Ablation results of Rose-SQL on the CoSQL dev set, illustrating the cumulative performance gains.}
    \label{tab:ablation_cosql}
\end{table}

\section{Ablation Study}
We conduct extensive ablation experiments on the SParC and CoSQL development sets to verify the effectiveness of each component in Rose-SQL and to examine how different forms of structural guidance contribute to robust multi-turn Text-to-SQL generation. To establish a rigorous in-context learning baseline, we use Qwen3-8B prompted with three random examples under standard CoT reasoning. We then incrementally introduce the Role-State guided reasoning path, Role Knowledge Injection ($\mathcal{E}_{rki}$), Evolutionary Trajectories ($\mathcal{P}_i$), and the Contextual Anchor ($\mathcal{C}_{ctx}$), with results reported in Table~\ref{tab:ablation_sparc} and Table~\ref{tab:ablation_cosql}.

\paragraph{Influence of Role-State Incorporation.}
As shown in the ablation results, replacing conventional chain-of-thought prompting with the Role-State guided reasoning path yields clear gains in both QM-EX and IM-EX, together with consistent improvements in TS. These results indicate that introducing a structural intermediate target effectively reduces the reasoning burden of multi-turn Text-to-SQL generation. Although the initial changes in QM-EM and IM-EM are relatively modest, the substantial improvements in execution-based metrics suggest that Role-State already helps the model capture the functional structure of the target SQL even before exact syntactic alignment is fully optimized. This validates the role of Role-State as a structural blueprint that decouples semantic intent from syntactic realization.

\paragraph{Impact of Role Knowledge Injection.}
After introducing Role Knowledge Injection ($\mathcal{E}_{rki}$), the model exhibits a marked improvement in EM-related metrics, especially on SParC. Specifically, QM-EM and IM-EM increase by 7.06\% and 8.93\%, respectively. This trend shows that atomic role-level exemplars provide useful prior knowledge about SQL compositional patterns and help the model translate abstract structural plans into syntactically precise SQL expressions. In other words, while Role-State mainly stabilizes the structural reasoning process, $\mathcal{E}_{rki}$ further strengthens the model's ability to instantiate that structure correctly at the SQL level.

\paragraph{Synergy of Evolutionary Trajectories and Contextual Anchors.}
The introduction of Evolutionary Trajectories ($\mathcal{P}_i$) and the Contextual Anchor ($\mathcal{C}_{ctx}$) further improves the overall performance, especially on interaction-level metrics. This indicates that their primary contribution lies in modeling dialogue evolution rather than only refining single-turn SQL structure. By aligning the current question with verified structural transitions and filtering out irrelevant historical noise, these two components jointly improve cross-turn consistency. The incremental gains observed after adding $\mathcal{P}_i$ and then $\mathcal{C}_{ctx}$ suggest that Rose-SQL benefits from a layered reasoning design: Role-State provides the structural foundation, while verified trajectories and contextual anchoring provide dynamic guidance for multi-turn reasoning.

\begin{table}[h]
    \centering
    \renewcommand{\arraystretch}{1.1}
    \setlength{\tabcolsep}{1.2mm}
    \begin{tabular}{c ccc ccc}
        \toprule
        \multirow{2}{*}{\textbf{$N$}} & \multicolumn{3}{c}{\textbf{QM}} & \multicolumn{3}{c}{\textbf{IM}} \\
        \cmidrule(lr){2-4} \cmidrule(lr){5-7}
        & \textbf{EM $\uparrow$} & \textbf{EX $\uparrow$} & \textbf{TS $\uparrow$} & \textbf{EM $\uparrow$} & \textbf{EX $\uparrow$} & \textbf{TS $\uparrow$} \\
        \midrule
        0 & 60.43 & 71.82 & 65.75 & 39.73 & 49.52 & 41.94 \\
        1 & 62.09 & 72.32 & 67.66 & 40.76 & 50.00 & 44.79 \\
        \textbf{2} & \textbf{65.75} & \textbf{78.72} & \textbf{72.73} & \textbf{50.00} & \textbf{62.56} & \textbf{55.21} \\
        3 & 63.51 & 75.39 & 70.32 & 41.70 & 53.31 & 47.87 \\
        4 & 64.09 & 74.48 & 69.91 & 44.79 & 54.74 & 48.58 \\
        5 & 63.51 & 74.06 & 69.33 & 42.89 & 53.08 & 48.10 \\
        \bottomrule
    \end{tabular}
    \caption{Impact of evolutionary trajectories count ($N$) on SParC dev set under the Rose-SQL (8B) setting.}
    \label{tab:NUM_Sparc}
\end{table}

\begin{table}[h]
    \centering
    \renewcommand{\arraystretch}{1.1}
    \setlength{\tabcolsep}{1.2mm}
    \begin{tabular}{c ccc ccc}
        \toprule
        \multirow{2}{*}{\textbf{$N$}} & \multicolumn{3}{c}{\textbf{QM}} & \multicolumn{3}{c}{\textbf{IM}} \\
        \cmidrule(lr){2-4} \cmidrule(lr){5-7}
        & \textbf{EM $\uparrow$} & \textbf{EX $\uparrow$} & \textbf{TS $\uparrow$} & \textbf{EM $\uparrow$} & \textbf{EX $\uparrow$} & \textbf{TS $\uparrow$} \\
        \midrule
        0 & 58.49 & 72.69 & 65.54 & 30.03 & 44.03 & 34.81 \\
        \textbf{1} & \textbf{63.16} & \textbf{74.48} & \textbf{68.62} & \textbf{36.18} & \textbf{50.85} & \textbf{41.30} \\
        2 & 61.37 & 74.28 & 68.22 & 33.11 & 48.12 & 39.59 \\
        3 & 60.97 & 73.98 & 67.92 & 32.42 & 48.46 & 39.93 \\
        4 & 61.37 & 73.68 & 68.12 & 35.84 & 46.76 & 39.93 \\
        5 & 60.38 & 73.49 & 68.02 & 32.08 & 47.10 & 39.59 \\
        \bottomrule
    \end{tabular}
    \caption{Impact of evolutionary trajectories count ($N$) on CoSQL dev set under the Rose-SQL (8B) setting.}
    \label{tab:NUM_Cosql}
\end{table}

\paragraph{Effect of Evolutionary Trajectory Quantity.}
To investigate the effect of trajectory guidance strength, we further vary the number of provided Evolutionary Trajectories ($N$). As shown in Table~\ref{tab:NUM_Sparc} and Table~\ref{tab:NUM_Cosql}, incorporating trajectory guidance consistently improves performance over the no-trajectory setting ($N=0$), confirming that retrieved structural references are beneficial for multi-turn SQL generation. On SParC, the best performance is achieved at $N=2$, where QM-EX and IM-EX improve by 6.40\% and 12.56\%, respectively, compared with the $N=1$ setting. On CoSQL, the optimal setting is $N=1$. When $N$ continues to increase, performance gradually declines on both benchmarks. This pattern suggests that trajectory guidance is beneficial only when the amount of retrieved structural evidence remains well controlled: too few candidates may provide insufficient support, whereas too many candidates introduce structural ambiguity and contextual noise, making it harder for the model to identify the most appropriate reasoning path.

\paragraph{Contribution of Contextual Anchoring.}
The final row of Table~\ref{tab:ablation_sparc} and Table~\ref{tab:ablation_cosql} highlights the importance of the Contextual Anchor ($\mathcal{C}_{ctx}$). By explicitly focusing the model on the most relevant historical reference, $\mathcal{C}_{ctx}$ improves both question-level and interaction-level generation quality. These gains confirm that, in multi-turn Text-to-SQL, accurate structural reasoning alone is not sufficient; the model must also remain grounded in the correct portion of the dialogue history. The contribution of $\mathcal{C}_{ctx}$ is therefore complementary to that of Role-State and trajectory retrieval, and together they form a coherent reasoning pipeline for context-dependent parsing.

\paragraph{Comparison with SFT Baselines.}
To further examine whether the gains of Rose-SQL arise mainly from model scale or from the proposed structured reasoning mechanism, we compare Qwen3 reasoning models against traditional supervised fine-tuned (SFT) baselines, as further detailed in Appendix \ref{Ablation Studies}. The comparison shows that weaker base reasoning models may underperform task-specific SFT experts under vanilla chain-of-thought prompting, whereas the introduction of Rose-SQL substantially narrows this gap and, in stronger settings, can even reverse it. These findings indicate that the efficacy of Rose-SQL does not simply arise from the scaling properties of the backbone model, but from the structural guidance introduced by the framework.

\begin{figure}[htbp]
    \centering
    \includegraphics[width=0.49\textwidth]{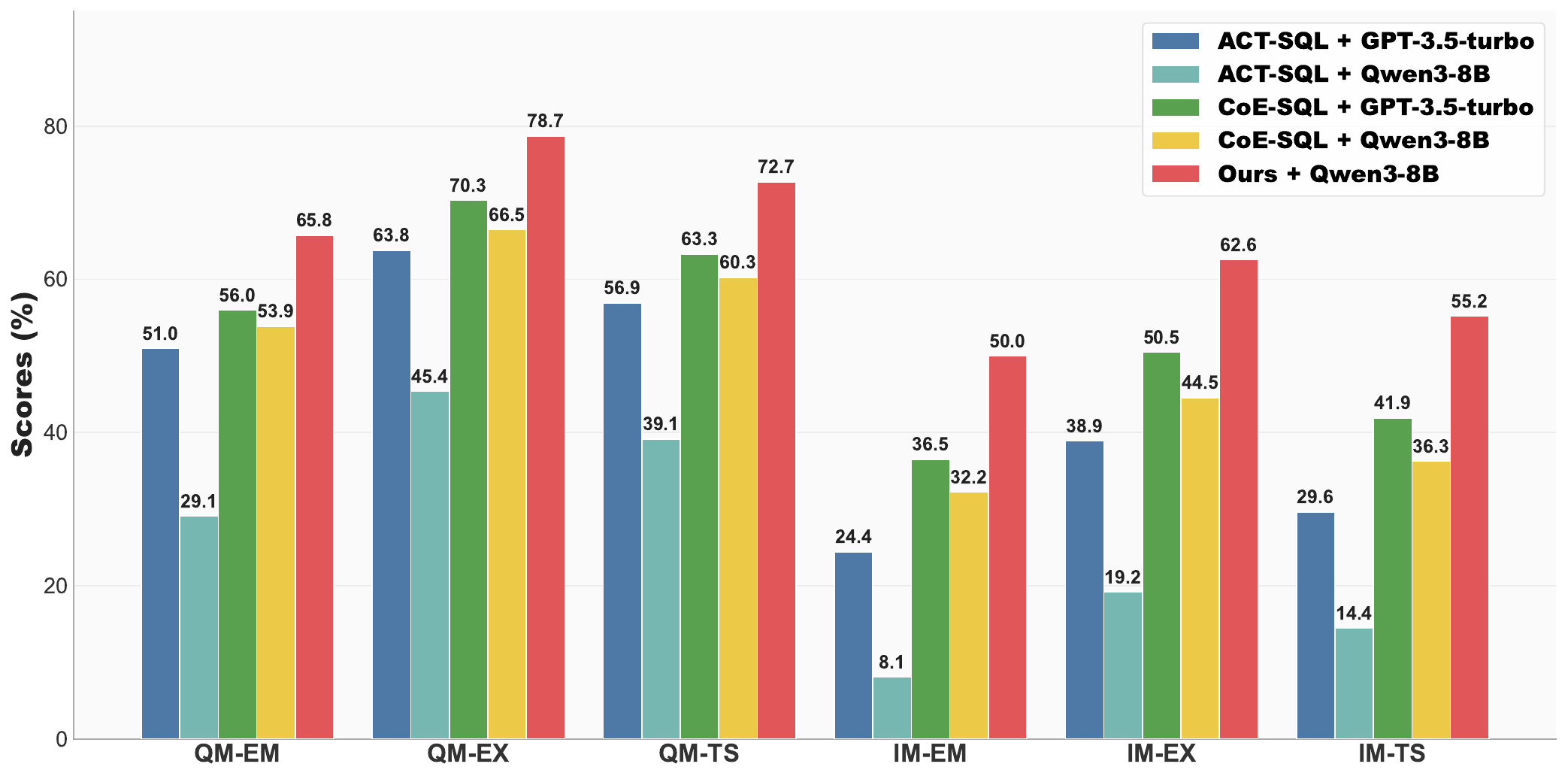}
    \caption{Performance comparison of Rose-SQL (8B) against established in-context learning baselines under the model replacement setting on the SParC development set.}
    \label{fig:icl_rose_sparc}
\end{figure}

\begin{figure}[htbp]
    \centering
    \includegraphics[width=0.49\textwidth]{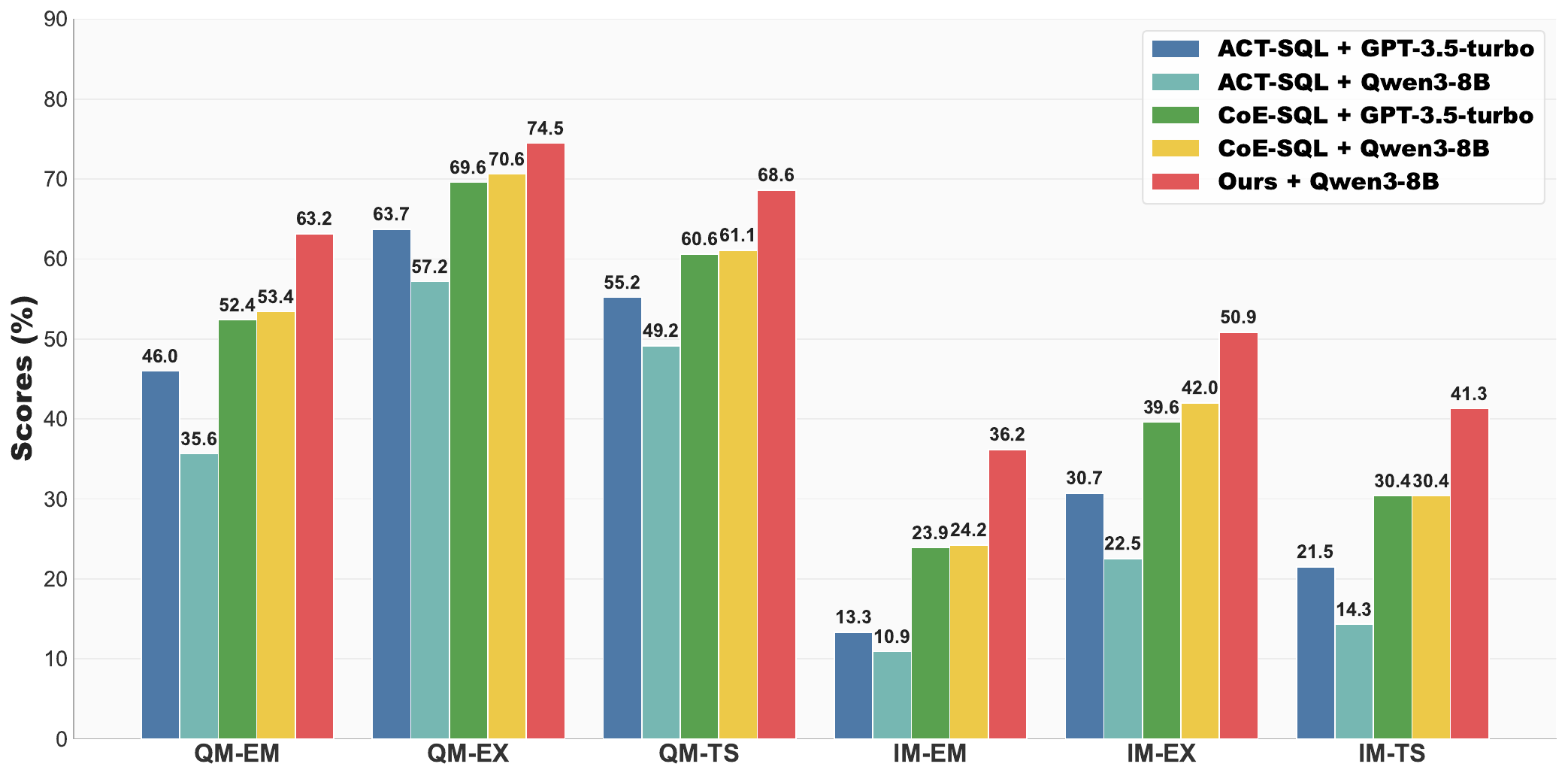}
    \caption{Performance comparison of Rose-SQL (8B) against established in-context learning baselines under the model replacement setting on the CoSQL development set.}
    \label{fig:icl_rose_cosql}
\end{figure}

\paragraph{Compatibility with Reasoning Models.}
We further assess the compatibility of Rose-SQL with reasoning-oriented backbones by comparing it against established in-context learning baselines under a model replacement setting. As illustrated in Figure~\ref{fig:icl_rose_sparc} and Figure~\ref{fig:icl_rose_cosql}, prior methods such as ACT-SQL exhibit substantial degradation when directly transferred to Qwen3-8B, with QM-EX and IM-EX on SParC declining by 18.41\% and 19.71\%, respectively. In contrast, Rose-SQL consistently maintains stronger performance across all metrics. This result suggests that Rose-SQL is better aligned with the reasoning behavior of small-scale LRMs: instead of relying mainly on prompt-sensitive demonstrations, it provides explicit structural guidance through Role-State and verified evolutionary trajectories, thereby enabling the model to utilize dialogue history more reliably in multi-turn semantic parsing.

\section{Related Work}
\paragraph{Large Reasoning Models.} Large Reasoning Models (LRMs) mark a significant advancement in LLMs' capabilities through the integration of explicit reasoning processes into LLMs. Unlike traditional LLMs that directly generate answers, LRMs like DeepSeek-R1 \cite{DeepSeek-R1}, OpenAI-O1 \cite{OpenAI-O1} and Qwen3 series models \cite{Qwen3} leverage advanced reasoning techniques, such as CoT \cite{CoT}, self-reflection \cite{self-refine} and roblem divide-andconquer \cite{tang2025,plaat, ToT}, to tackle complex problems in mathematics, code, and scientific domains \cite{ReasonningSurvey, Huang, protoReasoning}. For complex reasoning tasks, LRMs significantly outperform previous general-purpose LLMs like DeepSeek-V3 \cite{deepseek-v3} and GPT-4o \cite{GPT-4o}, demostrating the effectiveness of Long CoT training and specialized architectures.

\paragraph{Multi-Turn Text-to-SQL.} Research in this field can mainly be divided into two categories: In-context learning (ICL) methods and fine-tuning methods. ICL methods include  ACT-SQL \cite{ACT-SQL}, which removes the need to manually create CoT exemplars and automatically generating valuable demonstrations; CoE-SQL \cite{CoE-SQL} keeps track of user intentions by serializing modifications in SQL queries. Fine-tuning methods mainly focus on building specialized deep neural networks. IGSQL \cite{IGSQL} leverages graph neural networks to model database schema items within conversational contexts; R$^{2}$-SQL \cite{R^2SQL} and HIE-SQL \cite{HIE-SQL} propose a dynamic schema-linking graph that integrates the current utterance, preceding utterances, database schema, and the most recent SQL query; RASAT \cite{RASAT} harnesses relational self-attention mechanisms to capture the relationships between text and schemas; QDA-SQL employs LLMs to generate diverse multi-turn Text-to-SQL dialogues, training models to tackle ambiguous, unanswerable, and improper questions in conjunction with SQL queries; Track-SQL \cite{Track-SQL} conducts semantic enhancement and dynamic filtering to select the most relevant schema and historical SQL for the current query, thereby mitigating the impact of noisy context. Prior approaches either rely on large proprietary models sensitive to prompts quality or require costly fine-tuning of small models for context tracking. In contrast, Rose-SQL presents a training-free paradigm that directs small-scale reasoning models through verified Evolutionary Trajectories, facilitating precise context modeling without any specialized training overhead.

\section{Conclusion}

In this paper, we propose Rose-SQL, a training-free framework for multi-turn Text-to-SQL that addresses structural misalignment and contextual noise. The central idea of Rose-SQL is to introduce the Role-State as an explicit structural space for modeling dialogue evolution, so that the model reasons over verifiable structural transitions rather than relying solely on raw conversational history. Through its digitized structural form, the Role-State enables structural comparison across dialogue turns and supports the verification of evolutionary trajectories that provide reliable guidance for the current turn. Combined with augmented hierarchical reasoning, Rose-SQL offers a more stable path from multi-turn natural language intent to executable SQL generation. Experimental results on the SParC and CoSQL benchmarks demonstrate that Rose-SQL consistently improves the performance of small-scale reasoning models, highlighting the potential of training-free structural guidance for efficient context-dependent semantic parsing.

\section*{Limitations}

Rose-SQL achieves strong results on SParC and CoSQL, but three limitations remain: (1) Structural Retrieval Sparsity: structurally compatible trajectories can be scarce for rare dialogue patterns or specialized domains; (2) Role-State Expressiveness: the current lightweight Role-State may not fully cover richer open-domain SQL structures; (3) Inference Overhead: retrieved trajectories and explicit structural reasoning increase cost over direct prompting. Future work will improve sparse retrieval, relax exact structural matching, and make reasoning more efficient.

\section*{Acknowledgments}
This research was supported in part by National Science and Technology Major Project (2021ZD0111502), Natural Science Foundation of China (U24A20233, 62406078), CCF-DiDi GAIA Collaborative Research Funds (CCF-DiDi GAIA 202521), and Science and Technology Research and Development Plan of China Railway (P2025S001).

\section*{Ethics Statement}
We do not identify direct ethical concerns arising from this study. Our experiments are conducted solely on publicly available benchmarks released for academic research, and we do not collect or use private user data, personally identifiable information, or data involving human subjects. The base models and supporting toolkits used in our study are also publicly available for research use. In addition, our work focuses on an academic benchmark setting rather than real-world deployment.

\bibliography{custom}
\newpage

\appendix

\section*{Appendix}
\label{sec:appendix}

\begin{table*}[t]
    \centering
    \setlength{\tabcolsep}{1.2mm}
    \renewcommand{\arraystretch}{1.1} 

    \resizebox{\textwidth}{!}{%
    \begin{tabular}{c|ccc|ccc|ccc|ccc}
        \hline
        \multirowcell{3}{\textbf{Model}} & \multicolumn{6}{c|}{\textbf{SParC}} & \multicolumn{6}{c}{\textbf{CoSQL}} \\
        \cline{2-13}
        & \multicolumn{3}{c|}{\textbf{QM}} & \multicolumn{3}{c|}{\textbf{IM}} & \multicolumn{3}{c|}{\textbf{QM}} & \multicolumn{3}{c}{\textbf{IM}} \\
        \cline{2-13}
        & \textbf{EM $\uparrow$} & \textbf{EX $\uparrow$} & \textbf{TS $\uparrow$} & \textbf{EM $\uparrow$} & \textbf{EX $\uparrow$} & \textbf{TS $\uparrow$} & \textbf{EM $\uparrow$} & \textbf{EX $\uparrow$} & \textbf{TS $\uparrow$} & \textbf{EM $\uparrow$} & \textbf{EX $\uparrow$} & \textbf{TS $\uparrow$} \\
        \hline
        
        \multicolumn{13}{c}{\textbf{\textit{Supervised Fine-tuned Baselines}}} \\
        \hline
        SFT Codellama 7B & 59.18 & 67.99 & 61.51 & 36.96 & 46.68 & 39.33 & 50.54 & 60.17 & 54.51 & 17.74 & 28.66 & 23.54 \\
        SFT Deepseek 7B  & 64.33 & 71.40 & 65.08 & 43.36 & 50.71 & 43.36 & 54.71 & 66.03 & 58.88 & 23.20 & 34.12 & 26.96 \\
        SFT Mistral 7B   & 64.17 & 70.82 & 65.58 & 43.60 & 52.13 & 45.49 & 56.20 & 64.94 & 59.68 & 24.57 & 34.81 & 29.01 \\
        \hline

        \multicolumn{13}{c}{\textbf{\textit{Qwen3 Reasoning Models}}} \\
        \hline
        Qwen3-4B & 54.11 & 66.00 & 58.27 & 28.67 & 40.28 & 31.04 & 54.20 & 62.07 & 56.11 & 27.30 & 31.06 & 25.25 \\
        + Role-State    & 57.77 & 71.57 & 65.84 & 31.04 & 48.10 & 41.23 & 60.68 & 70.21 & 63.65 & 30.38 & 42.66 & 32.76 \\
        \cdashline{1-13}
        Qwen3-8B & 53.12 & 66.99 & 61.68 & 26.78 & 41.94 & 35.31 & 56.50 & 65.32 & 60.66 & 25.94 & 34.57 & 30.38 \\
        + Role-State    & 53.37 & 72.57 & 65.37 & 30.80 & 48.57 & 39.57 & 57.56 & 72.17 & 66.07 & 24.48 & 40.00 & 31.72 \\
        \cdashline{1-13}
        Qwen3-14B & 55.19 & 71.98 & 65.25 & 29.38 & 49.05 & 40.52 & 58.78 & 70.19 & 63.14 & 27.64 & 42.32 & 32.76 \\
        + Role-State     & 58.94 & 73.82 & 67.58 & 32.70 & 51.18 & 43.36 & 60.38 & 76.46 & 69.61 & 28.67 & 50.17 & 38.57 \\
        \hline
    \end{tabular}%
    }
    \caption{Detailed performance comparison between Qwen3 reasoning models and SFT baselines on SParC and CoSQL development sets.}
    \label{tab:lrm_vs_sft_detailed}
\end{table*}

\section{Detail Experiment Setup}
All experiments were conducted on a highperformance server configured with an NVIDIA A800-SXM4-80GB GPU, a AMD EPYC 7763 64-Core Processor, 2TB of memory, and the Ubuntu 22.04.5 LTS operating system. Leveraging the open-source nature of the Qwen3 series models, we performed all inference tasks using vLLM \cite{VLLM} framework on our local infrastructure. This approach provides full control over the experimental environment, in contrast to relying on external API-based services. For efficient and high-throughput inference, we utilized the \texttt{vLLM} framework.

In our generation configuration, we employed a greedy decoding strategy by setting the sampling temperature to 0. This deterministic approach is crucial for the Text-to-SQL task. It ensures the reproducibility of our results and, more importantly, eliminates the stochasticity inherent in higher-temperature sampling. Randomness can lead to the generation of syntactically malformed or semantically incorrect SQL queries, whereas our objective is to consistently produce the most accurate and valid query for a given natural language question. Therefore, a deterministic generation process is essential for reliable evaluation.

\section{Details of Role Knowledge Injection}
\label{rki}
Role Knowledge Injection ($\mathcal{E}_{rki}$) provides atomic grounding for critical SQL functional roles. To optimize the trade-off between semantic relevance and context length, we implement a schema-consistent dynamic selection strategy. All potential exemplars are retrieved exclusively from the \texttt{college\_2} database within the SParC training set. 

For each current test query, the framework performs a semantic similarity search across the knowledge pool, followed by a reranking process based on similarity scores. Based on these rankings, we select the most relevant QA pairs to satisfy the requirement that each of the ten SQL roles is supported by two distinct exemplars. 
To maximize context efficiency, queries encompassing multiple functional roles are prioritized as joint structural references. We strictly ensure that no individual QA pair is duplicated within the final prompt. By reusing a single schema definition across all injected knowledge, this strategy significantly compresses the prompt length, effectively mitigating performance degradation in long-context reasoning while maintaining high semantic alignment with the current task.

\section{More Experiments and Ablation Studies}
\label{Ablation Studies}
\subsection{Effectiveness Across Model Scales and Baselines}

To verify that the performance gains of Rose-SQL are primarily driven by our proposed structural guidance rather than model-specific artifacts or sheer parameter scale, we conduct a comprehensive evaluation across the Qwen3 series and compare them against traditional supervised fine-tuned (SFT) experts. 

\paragraph{Model-Scale-Agnostic Enhancement.}
As detailed in Table~\ref{tab:lrm_vs_sft_detailed}, the integration of \textit{Role-State} provides a consistent and significant performance uplift regardless of the underlying model's size. On both SParC and CoSQL datasets, the structural guidance improves all major metrics for 4B, 8B, and 14B models. For instance, when applied to Qwen3-4B on the SParC dataset, Role-State improves the QM-EX metric by 5.57\% and the IM-EX metric by 7.82\%. This consistent pattern confirms that Role-State fundamentally simplifies the reasoning task, enabling models of varying capacities to reliably bridge the semantic gap between natural language and SQL.

\paragraph{Comparison with SFT Experts.}
A granular comparison with SFT baselines further highlights the efficiency of our methodology. Baseline reasoning models (4B/8B) relying solely on vanilla CoT often underperform SFT counterparts like Deepseek-7B and Mistral-7B. For example, the Qwen3-4B baseline achieves a QM-EM of 54.11\% on SParC, lower than the 64.17\% of SFT Mistral-7B. However, the introduction of Role-State structural anchors enables the 4B-scale model to achieve performance parity with SOTA fine-tuned methods and allows the 8B-scale model to significantly surpass them. These results emphasize that explicit structural modeling via Role-State is a more efficient and robust alternative to intensive task-specific fine-tuning.

\subsection{Granular Ablation Analysis}

To further dissect the performance gains of Rose-SQL, we conduct a fine-grained ablation study focusing on query difficulty and interaction turns. This analysis reveals how each structural and evolutionary component contributes to robust SQL generation in complex, multi-turn environments.

\paragraph{Ablation by Query Difficulty.}
We categorize queries in the SParC and CoSQL development sets into four difficulty tiers—Easy, Medium, Hard, and Extra—following the standard Spider criteria. As shown in Table~\ref{tab:ablation_difficulty_sparc} and Table~\ref{tab:ablation_difficulty_cosql}, the performance of the Qwen3-8B baseline significantly degrades as complexity increases, particularly in the "Hard" and "Extra" categories, where structural reasoning is most critical.

The introduction of \textit{Role-State} guided reasoning consistently improves Execution Accuracy (EX) across all tiers by establishing a stable structural intermediate representation. Furthermore, the addition of \textit{Role Knowledge Injection} ($\mathcal{E}_{rki}$) notably bolsters Exact Match (EM) scores, demonstrating that explicit semantic role priors are essential for structural correctness. The complete Rose-SQL framework, incorporating \textit{Evolutionary Trajectories} ($\mathcal{P}_i$) and the \textit{Contextual Anchor} ($\mathcal{C}_{ctx}$), achieves the most substantial gains in the "Extra" category, elevating SParC EX from 47.8\% to 53.0\%. This underscores the synergy between structural grounding and evolutionary context tracking in resolving high-complexity queries.

\begin{table*}[t]
    \centering
    \setlength{\tabcolsep}{0.8mm}
    \renewcommand{\arraystretch}{1.05}
    \resizebox{0.96\textwidth}{!}{%
    \begin{tabular}{l|ccc|ccc|ccc|ccc}
        \hline
        \multirowcell{2}{\textbf{Method}} 
        & \multicolumn{3}{c|}{\textbf{Easy (483)}} 
        & \multicolumn{3}{c|}{\textbf{Medium (441)}} 
        & \multicolumn{3}{c|}{\textbf{Hard (145)}} 
        & \multicolumn{3}{c}{\textbf{Extra (134)}} \\
        \cline{2-13}
        & \textbf{EM} & \textbf{EX} & \textbf{TS}
        & \textbf{EM} & \textbf{EX} & \textbf{TS}
        & \textbf{EM} & \textbf{EX} & \textbf{TS}
        & \textbf{EM} & \textbf{EX} & \textbf{TS} \\
        \hline
        Qwen3-8B (Base) & 76.8 & 73.9 & 70.4 & 49.0 & 66.4 & 57.6 & 28.3 & 55.2 & 44.1 & 17.2 & 47.8 & 32.1 \\
        + Role-State & 80.1 & 83.6 & 82.0 & 55.3 & 73.2 & 66.9 & 25.5 & 53.1 & 43.4 & 20.1 & 42.5 & 28.4 \\
        + $\mathcal{E}_{rki}$ & 79.5 & 82.4 & 79.5 & 58.7 & 73.7 & 67.1 & 34.5 & 55.2 & 46.9 & 25.4 & 45.5 & 32.1 \\
        + $\mathcal{P}_i$ & 82.2 & 84.0 & 80.5 & 60.3 & 75.8 & 69.6 & 39.3 & 60.0 & 53.1 & 29.1 & 48.5 & 32.1 \\
        \textbf{+ $\mathcal{C}_{ctx}$ (Rose-SQL)} & \textbf{85.3} & \textbf{89.7} & \textbf{87.0} & \textbf{63.9} & \textbf{80.2} & \textbf{73.6} & \textbf{37.2} & \textbf{61.4} & \textbf{52.4} & \textbf{32.1} & \textbf{53.0} & \textbf{40.3} \\
        \hline
    \end{tabular}%
    }
    \caption{Ablation results of Rose-SQL on SparC by difficulty. Brackets () indicate sample counts.}
    \label{tab:ablation_difficulty_sparc}
\end{table*}

\begin{table*}[t]
    \centering
    \setlength{\tabcolsep}{0.7mm}
    \renewcommand{\arraystretch}{1.05}
    \resizebox{0.96\textwidth}{!}{%
    \begin{tabular}{l|ccc|ccc|ccc|ccc}
        \hline
        \multirowcell{2}{\textbf{Method}} 
        & \multicolumn{3}{c|}{\textbf{Easy (417)}} 
        & \multicolumn{3}{c|}{\textbf{Medium (320)}} 
        & \multicolumn{3}{c|}{\textbf{Hard (163)}} 
        & \multicolumn{3}{c}{\textbf{Extra (107)}} \\
        \cline{2-13}
        & \textbf{EM} & \textbf{EX} & \textbf{TS}
        & \textbf{EM} & \textbf{EX} & \textbf{TS}
        & \textbf{EM} & \textbf{EX} & \textbf{TS}
        & \textbf{EM} & \textbf{EX} & \textbf{TS} \\
        \hline
        Qwen3-8B (Base) & 81.1 & 70.7 & 69.1 & 52.2 & 61.6 & 53.1 & 35.0 & 58.9 & 49.1 & 13.1 & 34.6 & 25.2 \\
        + Role-State & 81.8 & 81.5 & 78.9 & 59.1 & 70.3 & 62.2 & 39.9 & 62.6 & 53.4 & 15.0 & 37.4 & 24.3 \\
        + $\mathcal{E}_{rki}$ & 83.0 & 85.6 & 81.8 & 58.4 & 73.4 & 67.5 & 35.6 & 61.3 & 50.9 & 16.8 & 37.4 & 28.0 \\
        + $\mathcal{P}_i$ & 82.9 & 87.1 & 84.2 & 57.8 & 74.5 & 65.0 & 39.9 & 63.8 & 52.1 & 18.7 & 40.2 & 26.2 \\
        \textbf{+ $\mathcal{C}_{ctx}$ (Rose-SQL)} & \textbf{84.4} & \textbf{88.2} & \textbf{86.3} & \textbf{58.8} & \textbf{75.3} & \textbf{67.5} & \textbf{46.6} & \textbf{64.4} & \textbf{54.0} & \textbf{24.3} & \textbf{43.0} & \textbf{29.9} \\
        \hline
    \end{tabular}%
    }
    \caption{Ablation results of Rose-SQL on CoSQL by difficulty. Brackets () indicate sample counts.}
    \label{tab:ablation_difficulty_cosql}
\end{table*}

\begin{table*}[htbp]
    \centering
    \setlength{\tabcolsep}{1.2mm}
    \renewcommand{\arraystretch}{1.1}
    \resizebox{\textwidth}{!}{%
    \begin{tabular}{l|ccc|ccc|ccc|ccc|ccc}
        \hline
        \multirowcell{2}{\textbf{Method}} & \multicolumn{3}{c|}{\textbf{Turn 1 (422)}} & \multicolumn{3}{c|}{\textbf{Turn 2 (422)}} & \multicolumn{3}{c|}{\textbf{Turn 3 (270)}} & \multicolumn{3}{c|}{\textbf{Turn 4 (88)}} & \multicolumn{3}{c}{\textbf{Turn $>$ 4 (1)}} \\
        \cline{2-16}
        & \textbf{EM} & \textbf{EX} & \textbf{TS} & \textbf{EM} & \textbf{EX} & \textbf{TS} & \textbf{EM} & \textbf{EX} & \textbf{TS} & \textbf{EM} & \textbf{EX} & \textbf{TS} & \textbf{EM} & \textbf{EX} & \textbf{TS} \\
        \hline
        Qwen3-8B (Base) & 69.4 & 73.2 & 69.9 & 53.1 & 64.7 & 55.5 & 38.1 & 56.7 & 45.9 & 35.2 & 65.9 & 53.4 & 0 & 100 & 100 \\
        + Role-State    & 71.8 & 80.3 & 77.0 & 55.9 & 69.2 & 62.3 & 44.1 & 65.2 & 57.4 & 42.0 & 60.2 & 54.5 & 0 & 100 & 100 \\
        + $\mathcal{E}_{rki}$ & 72.3 & 78.2 & 74.6 & 58.8 & 71.3 & 64.0 & 48.5 & 64.4 & 57.0 & 48.9 & 65.9 & 58.0 & 0 & 100 & 100 \\
        + $\mathcal{P}_i$     & 73.0 & 78.9 & 75.8 & 59.7 & 71.1 & 64.7 & 51.5 & 67.4 & 61.9 & 51.1 & 67.0 & 61.4 & 0 & 100 & 100 \\
        \textbf{+ $\mathcal{C}_{ctx}$ (Rose-SQL)} & \textbf{77.3} & \textbf{86.5} & \textbf{82.9} & \textbf{63.3} & \textbf{76.3} & \textbf{69.2} & \textbf{54.8} & \textbf{71.9} & \textbf{64.1} & \textbf{55.7} & \textbf{75.0} & \textbf{68.2} & \textbf{100} & \textbf{0} & \textbf{0} \\
        \hline
    \end{tabular}%
    }
    \caption{Ablation results of Rose-SQL on SParC dev set by turn. Brackets () indicate sample counts.}
    \label{tab:ablation_turns_sparc}
\end{table*}

\begin{table*}[h]
    \centering
    \setlength{\tabcolsep}{1.2mm}
    \renewcommand{\arraystretch}{1.1}
    \resizebox{\textwidth}{!}{%
    \begin{tabular}{l|ccc|ccc|ccc|ccc|ccc}
        \hline
        \multirowcell{2}{\textbf{Method}} & \multicolumn{3}{c|}{\textbf{Turn 1 (293)}} & \multicolumn{3}{c|}{\textbf{Turn 2 (285)}} & \multicolumn{3}{c|}{\textbf{Turn 3 (244)}} & \multicolumn{3}{c|}{\textbf{Turn 4 (114)}} & \multicolumn{3}{c}{\textbf{Turn $>$ 4 (71)}} \\
        \cline{2-16}
        & \textbf{EM} & \textbf{EX} & \textbf{TS} & \textbf{EM} & \textbf{EX} & \textbf{TS} & \textbf{EM} & \textbf{EX} & \textbf{TS} & \textbf{EM} & \textbf{EX} & \textbf{TS} & \textbf{EM} & \textbf{EX} & \textbf{TS} \\
        \hline
        Qwen3-8B (Base) & 60.1 & 70.0 & 64.2 & 57.2 & 63.5 & 56.5 & 59.4 & 59.4 & 53.3 & 52.6 & 56.1 & 49.1 & 45.1 & 42.3 & 42.3 \\
        + Role-State    & 64.8 & 73.7 & 68.3 & 58.6 & 71.2 & 64.6 & 62.3 & 71.7 & 63.5 & 57.9 & 67.5 & 58.8 & 50.7 & 50.7 & 49.3 \\
        + $\mathcal{E}_{rki}$ & 63.5 & 75.8 & 70.3 & 61.8 & 77.5 & 70.2 & 63.5 & 75.4 & 67.6 & 55.3 & 72.8 & 63.2 & 49.3 & 63.4 & 57.7 \\
        + $\mathcal{P}_i$     & 63.5 & 76.8 & 71.3 & 63.5 & 77.5 & 70.5 & 58.6 & 75.4 & 66.0 & 54.4 & 71.1 & 62.3 & 53.5 & 66.2 & 59.2 \\
        \textbf{+ $\mathcal{C}_{ctx}$ (Rose-SQL)} & \textbf{67.9} & \textbf{78.8} & \textbf{72.3} & \textbf{64.5} & \textbf{78.4} & \textbf{71.8} & \textbf{62.7} & \textbf{77.0} & \textbf{70.2} & \textbf{60.5} & \textbf{72.8} & \textbf{67.5} & \textbf{54.1} & \textbf{68.2} & \textbf{63.6} \\
        \hline
    \end{tabular}%
    }
    \caption{Ablation results of Rose-SQL on CoSQL dev set by turn. Brackets () indicate sample counts.}
    \label{tab:ablation_turns_cosql}
\end{table*}

\begin{table*}[!htbp]
    \centering
    \renewcommand{\arraystretch}{1.05}
    \setlength{\tabcolsep}{1.0mm}
    \resizebox{0.96\textwidth}{!}{%
    \begin{tabular}{lccccccc}
        \toprule
        \textbf{Dataset} & \textbf{Methods} & \textbf{Avg. Latency (s)} & \textbf{Throughput (q/s)} & \textbf{Avg. Input Tokens} & \textbf{Avg. Output Tokens} & \textbf{QM-EX (\%)} & \textbf{IM-EX (\%)} \\
        \midrule
        \multirow{4}{*}{SParC}
        & ACT-SQL (8B) & 11.26 & 0.09 & 7,559 & 866 & 45.39 & 19.19 \\
        & CoE-SQL (8B) & 14.64 & 0.07 & 6,569 & 1,142 & 66.50 & 44.55 \\
        & Qwen3-14B (Base) & 18.01 & 0.06 & 5,379 & 840 & 71.98 & 49.05 \\
        & \textbf{Rose-SQL (8B)} & \textbf{19.61} & \textbf{0.05} & \textbf{17,305} & \textbf{1,581} & \textbf{78.72} & \textbf{62.56} \\
        \midrule
        \multirow{4}{*}{CoSQL}
        & ACT-SQL (8B) & 10.84 & 0.09 & 7,214 & 841 & 57.19 & 22.53 \\
        & CoE-SQL (8B) & 13.58 & 0.07 & 6,214 & 1,084 & 70.61 & 41.98 \\
        & Qwen3-14B (Base) & 17.24 & 0.06 & 5,104 & 791 & 70.19 & 42.32 \\
        & \textbf{Rose-SQL (8B)} & \textbf{17.86} & \textbf{0.06} & \textbf{14,862} & \textbf{1,436} & \textbf{74.48} & \textbf{50.85} \\
        \bottomrule
    \end{tabular}%
    }
    \caption{Efficiency comparison on SParC and CoSQL under the same deterministic decoding configuration. }
    \label{tab:efficiency_comparison_both}
\end{table*}

\paragraph{Ablation by Interaction Turn.}
To evaluate Rose-SQL's ability to handle evolving dialogue history, we analyze performance across sequential turns (Table~\ref{tab:ablation_turns_sparc} and Table~\ref{tab:ablation_turns_cosql}). Baseline reasoning models typically suffer from contextual degradation as interactions progress, struggling to filter historical noise. 

Rose-SQL stabilizes this trend by maintaining conversational coherence through vectorized Role-State transitions. Crucially, the performance gap between Rose-SQL and the baseline widens in later turns—for instance, achieving a 75.0\% EX in the fourth turn on SParC compared to the baseline's 65.9\%. This demonstrates that \textit{Evolutionary Trajectory Searching} effectively prevents error propagation, ensuring that small-scale reasoning models remain precisely focused on relevant dependencies throughout long-range interactions.

\subsection{Analysis of Gain Dependency Components}
\begin{table}[!htbp]
    \centering
    \renewcommand{\arraystretch}{1.05}
    \setlength{\tabcolsep}{1.0mm}
    \resizebox{\columnwidth}{!}{%
    \begin{tabular}{lcccc}
        \toprule
        \multirow{2}{*}{\textbf{GDA Component}} & \multicolumn{2}{c}{\textbf{SParC}} & \multicolumn{2}{c}{\textbf{CoSQL}} \\
        \cmidrule(lr){2-3} \cmidrule(lr){4-5}
        & \textbf{QM-EX (\%)} & \textbf{IM-EX (\%)} & \textbf{QM-EX (\%)} & \textbf{IM-EX (\%)} \\
        \midrule
        Baseline (No GDA) & 72.57 & 48.57 & 72.17 & 40.00 \\
        Distance Only ($\mathbf{D}_i$) & 72.62 & 49.85 & 72.24 & 41.10 \\
        Strength Only ($\mathbf{S}_i$) & 76.50 & 58.42 & 73.68 & 47.60 \\
        \textbf{Full GDA ($\mathbf{S}_i \cdot \mathbf{D}_i$)} & \textbf{78.72} & \textbf{62.56} & \textbf{74.48} & \textbf{50.85} \\
        \bottomrule
    \end{tabular}%
    }
    \caption{Ablation of GDA components on SParC and CoSQL dev sets.}
    \label{tab:gda_components_both}
\end{table}
To further clarify the role of Gain Dependency Analysis (GDA), we isolate the contributions of Dependency Strength ($\mathbf{S}_i$) and Dependency Distance ($\mathbf{D}_i$) on both SParC and CoSQL. Specifically, we compare the full GDA design against three variants: a baseline without GDA, a distance-only variant using temporal proximity alone, and a strength-only variant using perplexity-based dependency strength alone. The results are reported in Table~\ref{tab:gda_components_both}.

As shown in Table~\ref{tab:gda_components_both}, temporal distance alone brings only marginal improvement over the no-GDA baseline, indicating that recency by itself is insufficient for identifying informative dialogue history. In contrast, dependency strength yields a much larger gain, confirming that perplexity reduction captures the main signal of useful historical dependence. The full GDA formulation, which combines dependency strength and temporal distance, achieves the best overall performance. This result suggests that dependency strength provides the dominant signal, while temporal distance further helps suppress semantically similar but contextually less relevant turns.

\section{Efficiency Analysis}

To quantify the computational overhead introduced by Rose-SQL, we evaluate inference efficiency under the same hardware environment and deterministic decoding configuration. We compare Rose-SQL against representative prompting-based baselines and a larger base reasoning model.

As shown in Table~\ref{tab:efficiency_comparison_both}, Rose-SQL introduces higher latency and token usage than standard 8B prompting baselines, mainly because the prompt additionally includes retrieved trajectories, contextual anchors, and role-level exemplars. However, its runtime remains comparable to directly deploying a larger 14B base model, while achieving substantially stronger multi-turn execution accuracy. This reflects a practical trade-off of Rose-SQL: it exchanges context length for parameter scale, allowing a training-free 8B model to approach or surpass stronger baselines without task-specific fine-tuning. Importantly, the added overhead comes primarily from prompt composition rather than larger model parameters, and inference is still completed in a single deterministic decoding pass.

\section{Prompts Used in Our Methodology}
\label{sec:appendix_prompts}

In this section, we present the prompt templates utilized for invoking the Large Reasoning Models (LRMs) within the Rose-SQL framework. Our methodology involves two primary stages: (1) the offline generation of \textit{Role-State Reasoning} for training data construction, and (2) the online generation of SQL queries during the inference phase.

The prompt template used to instruct the LRM to deduce the reasoning process for the training set is presented in Table~\ref{tab:prompt_rolestate_construction}. This template is structured in an agent-like format to ensure the model strictly adheres to the schema-linking logic and the structural constraints of the \textit{Role-State}. For the inference phase, the prompt template presented in Table~\ref{tab:prompt_inference} incorporates few-shot exemplars and historical dialogue context, guiding the model to perform stable and coherent multi-turn reasoning.

\begin{table*}[htbp]
\centering
\begin{tabular}{p{1\linewidth}}
\hline
\textbf{System}:\\
You are an NL2SQL Research Expert. Following is some information about your profile and task constraints:

\#\# Profile\\
- **Task Objective**: Given a database schema, a question (Q), predefined intermediate representation (Roles), and the corresponding SQL, deduce and articulate the complete thought process.\\
- **Core Methodology**: You must bridge the gap from Q to SQL via the \textit{Role-State} intermediate representation, ensuring logical consistency and structural grounding.

\#\# Workflows\\
1. **Schema Linking**: Integrate database schema information with question Q to deduce and populate each part of Roles.\\
2. **SQL Generation**: Generate the final SQL statement based on the Roles representation.\\
3. **Validation**: Ensure all referenced database objects strictly match the provided schema and the SQL query is on a single line.

\#\# Rules\\
1. **Data Consistency**: All referenced objects must strictly match the schema (case-sensitive). Aliases (AS) are prohibited.\\
2. **SQL Specifications**: Use lowercase throughout (except string constants). Final SQL must be on a single line.\\
3. **Ambiguity Handling**: If schema ambiguity exists, document assumptions and proceed with reasonable guesses.\\
4. **Context Handling**: Critically evaluate historical context. The context is a tool, not a mandate; do not blindly follow previous turn logic if independent.

\#\# Output Format\\
Strictly output in the following format:
\begin{itemize}
    \item \texttt{<step>Step 1: ...</step>} (Detailed deduction of Roles and SQL)
    \item \texttt{<Roles>Copy provided Roles verbatim</Roles>}
    \item \texttt{<SQL>Copy provided SQL verbatim</SQL>}
\end{itemize}

\textbf{User}:\\
"question": "\{question\}", \\
"database": "\{database\}", \\
"Roles": "\{Roles\}", \\
"SQL": "\{SQL\}", \\

The process of reasoning:\\
\hline
\end{tabular}
\caption{Prompt template used for generating the \textit{Role-State Reasoning} for training set construction.}
\label{tab:prompt_rolestate_construction}
\end{table*}

\begin{table*}[htbp]
\centering
\begin{tabular}{p{1\linewidth}}
\hline
\textbf{System}:\\
You are an NL2SQL Reasoning Expert. Your task is to act as a database and logic expert to convert a natural language question into a precise SQL query.

\#\# Profile\\
- **Objective**: Apply step-by-step reasoning to produce a full thought process (Reasoning, Roles) and the final SQL query for the current question.\\
- **Capabilities**: Expert in schema mapping, logical deduction, and multi-turn dialogue management.

\#\# Workflows\\
1. **Analogy Learning**: Review provided exemplars to understand the step-by-step reasoning methodology.\\
2. **Context Analysis**: Evaluate previous questions and reasoning (if provided) to identify relevant historical dependencies.\\
3. **Hierarchical Reasoning**: Conduct Schema Linking to populate Roles, then generate the SQL query.\\
4. **Refinement**: Verify that the SQL is lowercase, uses correct join logic, and matches string filtering requirements (LIKE vs. =).

\#\# Rules\\
1. **Inference Logic**: Avoid repeated incorrect thinking; conduct efficient reasoning to obtain the \texttt{<SQL>} block.\\
2. **Roles Specifications**: Items under \textit{join/condition/group/order} must be in \texttt{table.column} form. For multi-turn transitions, \textit{union/except/intersect/in/nin} must be \textit{Exist} or \textit{None}.\\
3. **SQL Constraints**: No \texttt{table.*} expressions. For single tables, use \texttt{table.count(*)} in Roles for \texttt{SELECT COUNT(*)}. No \texttt{AS} keyword allowed.\\
4. **Historical Evaluation**: Evaluate relevance of previous turns. Do not follow previous logic if the current turn introduces independent constraints.

\#\# Output Format\\
Strictly output in the following order:
\begin{itemize}
    \item \texttt{<step>Step 1: ...</step>} (N steps)
    \item \texttt{<Roles>Constructed Role-State</Roles>}
    \item \texttt{<SQL>Single-line SQL query</SQL>}
\end{itemize}

\textbf{User}:\\
"exemplars": "\{exemplars\}", \\
"previous question": "\{pre\_question\}", \\
"previous reasoning": "\{pre\_rser\}", \\
"question": "\{question\}", \\
"database": "\{database\}", \\

Now, begin your analysis for the current task:\\
\hline
\end{tabular}
\caption{Prompt template used for the final SQL query generation (Inference phase).}
\label{tab:prompt_inference}
\end{table*}

\begin{table*}[t]
\centering
\small
\renewcommand{\arraystretch}{1.2}
\setlength{\dashlinedash}{0.5pt}
\setlength{\dashlinegap}{1.5pt}

\begin{tabular}{>{\centering\arraybackslash}m{1.2cm}|>{\centering\arraybackslash}m{1.8cm}|p{10.2cm}}
\hline
\multirow{7}{*}{Case \#1} 
& Question \#1 & What is every student's id? \\
& Question \#2 & Of those ids, which correspond to those who own cats as pets? \\
& Question \#3 & List all the other ids. \\
\cdashline{2-3}
& Gold & \textbf{SELECT stuid FROM student EXCEPT SELECT T1.stuid FROM student AS T1 JOIN has\_pet AS T2 ON T1.stuid = T2.stuid JOIN pets AS T3 ON T3.petid = T2.petid WHERE T3.pettype = `cat'} \\
\cdashline{2-3}
& Rose-SQL & \textcolor{teal}{SELECT stuid FROM student EXCEPT SELECT T1.stuid FROM student AS T1 JOIN has\_pet AS T2 ON T1.stuid = T2.stuid JOIN pets AS T3 ON T3.petid = T2.petid WHERE T3.pettype = `cat'} \\
\cdashline{2-3}
& w/o Traj. & SELECT stuid FROM has\_pet \textcolor{orange}{INTERSECT} SELECT stuid FROM has\_pet JOIN pets ON has\_pet.petid = pets.petid WHERE pettype = `cat' \\
\cdashline{2-3}
& Base LRM & SELECT student.stuid FROM student JOIN has\_pet ON student.stuid = has\_pet.stuid JOIN pets ON has\_pet.petid = pets.petid WHERE pets.pettype = `cat' \\
\hline
\multirow{6}{*}{Case \#2} 
& Question \#1 & How many car models are produced in total? \\
& Question \#2 & How many in Germany? \\
& Question \#3 & How about in Japan? \\
\cdashline{2-3}
& Gold & \textbf{SELECT count(*) FROM MODEL\_LIST AS T1 JOIN CAR\_MAKERS AS T2 ON T1.Maker = T2.Id JOIN COUNTRIES AS T3 ON T2.Country = T3.CountryId WHERE T3.CountryName = `japan'} \\
\cdashline{2-3}
& Rose-SQL & \textcolor{teal}{SELECT COUNT( * ) FROM model\_list JOIN car\_makers ON model\_list.maker = car\_makers.id JOIN countries ON car\_makers.country = countries.countryid WHERE countries.countryname = `japan'} \\
\cdashline{2-3}
& w/o Anchor & SELECT COUNT( * ) FROM model\_list JOIN car\_makers ON model\_list.maker = car\_makers.id \textcolor{orange}{WHERE countries.countryname = `germany'} \\
\cdashline{2-3}
& Base LRM & SELECT COUNT( * ) FROM model\_list \\
\hline
\multirow{4}{*}{Case \#3} 
& Question & What are all the nations? \\
\cdashline{2-3}
& Gold & \textbf{SELECT * FROM country} \\
\cdashline{2-3}
& Rose-SQL & \textcolor{teal}{SELECT *} FROM country \\
\cdashline{2-3}
& w/o Role-State & SELECT \textcolor{orange}{name} FROM country \\
\hline
\multirow{4}{*}{Case \#4} 
& Question & Show all the available features. \\
\cdashline{2-3}
& Gold & \textbf{SELECT * FROM Other\_Available\_Features} \\
\cdashline{2-3}
& Rose-SQL & \textcolor{teal}{SELECT *} FROM Other\_Available\_Features \\
\cdashline{2-3}
& w/o Role-State & SELECT \textcolor{orange}{feature\_name} FROM other\_available\_features \\
\hline
\multirow{5}{*}{Case \#5} 
& Question & Which continents have an average life expectancy less than age 72? \\
\cdashline{2-3}
& Gold & \textbf{SELECT sum(Population) , avg(LifeExpectancy), Continent FROM country GROUP BY Continent HAVING avg(LifeExpectancy) < 72} \\
\cdashline{2-3}
& Rose-SQL & \textcolor{teal}{SELECT sum(Population) , avg(LifeExpectancy), Continent} FROM country GROUP BY Continent HAVING avg(LifeExpectancy) < 72 \\
\cdashline{2-3}
& w/o Role-State & SELECT \textcolor{orange}{avg(lifeexpectancy), sum(population)} FROM country GROUP BY continent HAVING avg(lifeexpectancy) < 72 \\
\hline
\multirow{5}{*}{Case \#6} 
& Question \#1 & Tell me the director of the cartoon named Day of the Dark Knight!. \\
& Question \#2 & What is the channel of this cartoon? \\
\cdashline{2-3}
& Gold & \textbf{SELECT Channel FROM Cartoon WHERE Title = `Day of the Dark Knight!'} \\
\cdashline{2-3}
& Rose-SQL & SELECT \textcolor{teal}{channel} FROM cartoon WHERE Title = `Day of the Dark Knight!' \\
\cdashline{2-3}
& w/o Anchor & SELECT \textcolor{orange}{tv\_channel.series\_name} FROM \textcolor{orange}{cartoon JOIN tv\_channel ON cartoon.channel = tv\_channel.id} WHERE cartoon.title = `Day of the Dark Knight!' \\
\hline
\end{tabular}
\vspace{0.3cm}
\caption{Case study of Rose-SQL on the SParC dataset, comparing the full framework with various ablation configurations and the Base LRM. The color scheme is defined as follows: teal highlights accurate SQL structures and operators successfully generated by Rose-SQL, whereas orange signifies representative errors in alternative versions, including logical operator confusion (Case 1), conversational noise interference (Case 2, 6), and structural misalignment (Case 3, 4, 5).}
\label{tab:case_study}
\end{table*}

\section{Case Study}
To qualitatively evaluate the effectiveness of Rose-SQL, we present several representative cases in Table \ref{tab:case_study}, comparing the full framework against various ablation configurations and the Base LRM. 

First, Case 1 demonstrates the critical role of \textbf{Evolutionary Trajectories} $\mathcal{P}_i$ in capturing complex dialogue dynamics. Without this trajectory guidance (w/o Traj.), the model lacks the structural memory required to identify the transition toward subtractive logic in the final turn, erroneously employing an \textit{INTERSECT} operator instead of \textit{EXCEPT}. In contrast, Rose-SQL retrieves analogous vectorized transitions to enforce the correct logical structure, ensuring consistency across interactions.

Second, Cases 3, 4, and 5 illustrate the grounding effect of the \textbf{Role-State} as a structural blueprint. In the absence of this constraint (w/o Role-State), the model tends to hallucinate specific column names such as \textit{name} or \textit{feature\_name} when the intent requires a global selection \textit{SELECT *}. Furthermore, as shown in Case 5, the Role-State ensures that the \textit{GROUP BY} column is correctly reflected in the projection list, thereby maintaining the structural integrity of the generated query.

Finally, Cases 2 and 6 demonstrate the capacity of \textbf{Contextual Anchors} $\mathcal{C}_{ctx}$ to mitigate conversational noise. In Case 2, the version without anchors fails to update the country filter from Germany to Japan, a typical state-tracking error. In Case 6, the model introduces redundant table joins due to a lack of precise historical grounding. By utilizing anchors to isolate relevant turn-level information, Rose-SQL maintains high conversational coherence and structural precision even when utilizing small-scale reasoning models.

\end{document}